%% file: aaai24.tex
\documentclass[letterpaper]{article} % DO NOT CHANGE THIS
\usepackage{aaai24}  % DO NOT CHANGE THIS
\usepackage{times}  % DO NOT CHANGE THIS
\usepackage{helvet}  % DO NOT CHANGE THIS
\usepackage{courier}  % DO NOT CHANGE THIS
\usepackage[hyphens]{url}  % DO NOT CHANGE THIS
\usepackage{graphicx} % DO NOT CHANGE THIS
\usepackage{natbib}  % DO NOT CHANGE THIS AND DO NOT ADD ANY OPTIONS TO IT
\usepackage{caption} % DO NOT CHANGE THIS AND DO NOT ADD ANY OPTIONS TO IT
\usepackage{algorithm}
\usepackage{algpseudocode}

\usepackage{amsfonts}
\usepackage{amsmath}
\usepackage{multirow}
\usepackage{booktabs}
\usepackage{array}
\usepackage{graphicx}

\usepackage{color, colortbl}
\definecolor{Green}{rgb}{0.93,1,0.93}
\definecolor{Gray}{rgb}{0.85,0.85,0.85}
\usepackage{newfloat}
\usepackage{listings}
\usepackage{anyfontsize}
\DeclareCaptionStyle{ruled}{labelfont=normalfont,labelsep=colon,strut=off} % DO NOT CHANGE THIS
\floatstyle{ruled}
\newfloat{listing}{tb}{lst}{}
\floatname{listing}{Listing}
\usepackage{xcolor}

\newcommand{\sig}[0]{$^\ddagger$}
\definecolor{darkpink}{rgb}{0.96,0.14,0.56}
\usepackage{subcaption}
\usepackage{tabularx}
\title{READ: Recurrent Adapter with Partial Video-Language \\ Alignment for Parameter-Efficient  Transfer Learning in Low-Resource Video-Language Modeling}
\author{
    Thong Nguyen\textsuperscript{\rm 1}, \;
    Xiaobao Wu\textsuperscript{\rm 2}, \;
    Xinshuai Dong\textsuperscript{\rm 3}, \;
    Khoi Le\textsuperscript{\rm 4}, \;
    Zhiyuan Hu\textsuperscript{\rm 1}, \; \\
    Cong-Duy Nguyen\textsuperscript{\rm 2}, \;
    See-Kiong Ng\textsuperscript{\rm 1}, \;
    Anh Tuan Luu\textsuperscript{\rm 2}
}
\affiliations{
    \textsuperscript{\rm 1} Institute of Data Science (IDS), National University of Singapore, Singapore \\
    \textsuperscript{\rm 2} Nanyang Technological University (NTU), Singapore \\
    \textsuperscript{\rm 3} Carnegie Mellon University, USA \\
    \textsuperscript{\rm 4} VinAI Research, Vietnam \\
    \textcolor{darkpink}{thong.nguyen@u.nus.edu, seekiong@nus.edu.sg, anhtuan.luu@ntu.edu.sg}
}

\usepackage{bibentry}
\begin{document}

\maketitle

\begin{abstract}
% Finetuning a pretrained multi-modal transformer is becoming a new paradigm for video-language modeling tasks, such as temporal language grounding and video-language summarization.
% %
% As high-quality downstream data is often limited, training an adapter for each down-stream task often provides better training stability and parameter efficiency, compared to finetuning the whole model.
% %
% However, existing adapters fail to fully capture and utilize the fundamental characteristic of video-language modeling data: there is an intrinsic temporal dependency between video frames and textual words.  
% %
% As such, in this paper 
%  we introduce a novel lightweight adaptation framework, READ.
%  %
%  Specifically, to explicitly model temporal relations we propose a novel adaptor architecture, \textbf{RE}current \textbf{AD}apter (READ), that employs recurrent computation to enable temporal modeling capability.
%  %
%  Moreover, we propose a novel regularization term for training adapters, \textbf{P}artial \textbf{V}ideo-\textbf{L}anguage \textbf{A}lignment (PVLA), which facilitates maintaining inter-modal information in an adaptation later by minimizing partial optimal transport distance. 
%  %
%  We validate our READ framework through extensive experiments where READ significantly outperforms all existing fine-tuning strategies on multiple low-resource temporal language grounding and video-language summarization benchmarks.

Fully fine-tuning pretrained large-scale transformer models has become a popular paradigm for video-language modeling tasks, such as temporal language grounding and video-language summarization. With a growing number of tasks and limited training data, such full fine-tuning approach leads to costly model storage and unstable training. To overcome these shortcomings, we introduce lightweight adapters to the pre-trained model and only update them at fine-tuning time. However, existing adapters fail to capture intrinsic temporal relations among video frames or textual words. Moreover, they neglect the preservation of critical task-related information that flows from the raw video-language input into the adapter’s low-dimensional space. To address these issues, we first propose a novel \textbf{RE}current \textbf{AD}apter (READ) that employs recurrent computation to enable temporal modeling capability. Second, we propose \textbf{P}artial \textbf{V}ideo-\textbf{L}anguage \textbf{A}lignment (PVLA) objective via the use of partial optimal transport to maintain task-related information flowing into our READ modules. We validate our READ framework through extensive experiments where READ significantly outperforms all existing fine-tuning strategies on multiple low-resource temporal language grounding and video-language summarization benchmarks. The code, model, and data have been made available at \textcolor{darkpink}{nguyentthong.github.io/READ}.
\end{abstract}

\input{files/01_introduction}
\input{files/02_related_work}
\input{files/03_methodology}
\input{files/04_experiments}

\input{files/05_conclusion}
\input{files/06_ack}

\bibliography{aaai24}
% \appendix
% \input{files/appendix}

\end{document}

%% file: files/01_introduction.tex
\section{Introduction}
\begin{figure*}
    \centering
    \includegraphics[width=0.7\linewidth]{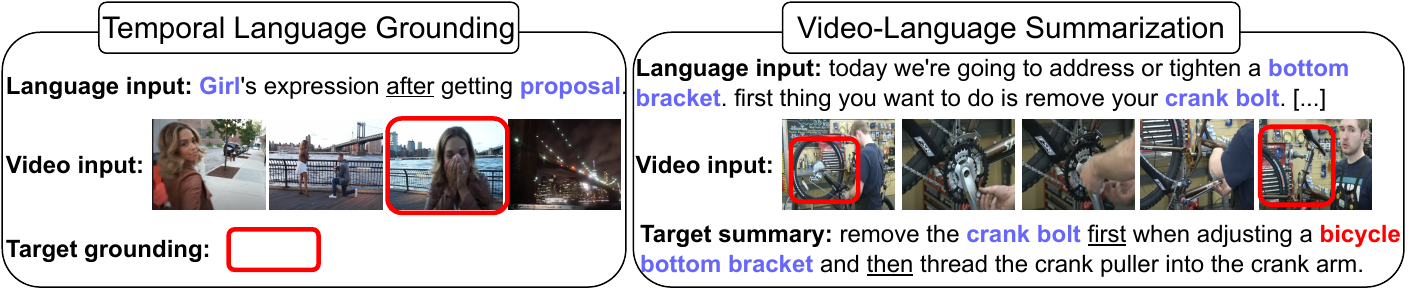}
    \caption{Examples of the TLG and VLS problems. TLG model needs to understand the meaning of language entities such as \textit{proposal} or \textit{girl}, and the existence of \textit{expression} in video frames. VLS model is expected to recognize salient information, \textit{e.g. crank bolt}, \textit{bottom bracket} from the language, and \textit{bicycle} from the video.}
    \label{fig:task_example}
    \vspace{-10pt}
\end{figure*}

Video-language modeling is a challenging problem since it involves understanding both video and language modalities. For example, temporal language grounding (TLG) model comprehends video detail and language query to localize semantically related video moments (Figure \ref{fig:task_example} (left)), or video-language summarization (VLS) model extracts information from both video content and language transcript to write the summary (Figure \ref{fig:task_example} (right)). 

Previous  video-language modeling methods \citep{liu2022umt, lei2021detecting, yu2021vision}  employ pretrained Transformer models such as Unified Multimodal Transformer (UMT) \citep{liu2022umt} and Vision-Guided BART (VG-BART) \citep{yu2021vision}, and fine-tune all the parameters of these models for every single task.
This results in substantial storage overhead since each task demands storing a separate model \citep{zhang2023multimodal}. Moreover, because of the difficulty of collecting video-language data \citep{pan2022st}, fully fine-tuning these over-parameterized models in low-resource scenarios, where limited training data is available, leads to instability and sub-optimal performance \citep{jiang2022cross, huang2023vop}.

% Video-language modeling is a challenging problem since it requires fine-grained multimodal understanding to extract semantically-related content \citep{gao2021relation} (\emph{e.g.} video-moment retrieval \citep{liu2022umt, gao2021relation, lei2021detecting}) or generate concise summary of video-language data (e.g. video-language summarization \citep{yu2021vision}), see Figure \ref{fig:task_example}. \wxb{Consider introducing an example task with video-language modeling.} Recently, multimodal large-scale Transformer models have emerged as an effective architecture for video-language modeling. \wxb{To fit downstream tasks, one common approach is to fully fine-tune a pre-trained Transformer model on video-language datasets.}
% % adapt a pre-trained Transformer model to downstream tasks by fully fine-tuning on video-language datasets.
% \wxb{However, this approach hinders model applicability because fitting multiple downstream tasks leads to expensive storage cost.}
% However, in the presence of multiple downstream tasks, obtaining a separate fine-tuned model for each task results in substantial storage cost and hinders the model applicability in real-world applications.
% In addition,
% \wxb{because of the difficulty of collecting video-language data and their limited scales \citep{pan2022st},}
% because video-language data is difficult to collect and often available in limited scale \citep{pan2022st}, fine-tuning these overparameterized models in low-resource scenarios \wxb{leads to} instability and sub-optimal performance \citep{jiang2022cross, huang2023vop}.

To address these shortcomings, adapters are proposed as a parameter-efficient solution for finetuning video-language pretrained transformers \citep{jiang2022cross, zhang2023multimodal, yang2023aim, sung2022vl, chen2022adaptformer}. 
The strategy is to add additional adaptation module to each layer of the pre-trained network and only the adaptation modules are trained during fine-tuning to improve the 
 parameter-performance trade-off.
These modules rely on non-linear projections to downproject video-language inputs into low-dimensional space then up-project them back to the original high-dimensional space.
However, such projections consider video frames and textual words as separate tokens, thus ignoring the intrinsic temporal dependency among video frames or textual words. Without such dependency information, it is difficult to reason about temporal context in the video to properly ground the language (\emph{e.g.} in Figure \ref{fig:task_example}, determine the \emph{expression} of the \emph{girl} \emph{after}, not \emph{before}, the \emph{proposal}), or coherently link the entities in the summary (\emph{e.g.} in Figure \ref{fig:task_example}, recap the chronological order of \emph{bolt removing} and \emph{puller threading}). Moreover, because at fine-tuning time only adaptation modules are trained using limited video-language data, little attention is paid to the information flow that starts from the raw video-language inputs till the low-dimensional space of the adaptation modules. This may result in losing essential task-related information and carrying noise into these modules \citep{tsai2020multimodal, han2021improving}. 
% In practice, both problems can worsen due to limited training data.

\begin{figure}[t]
    \centering
    \includegraphics[width=0.7\linewidth]{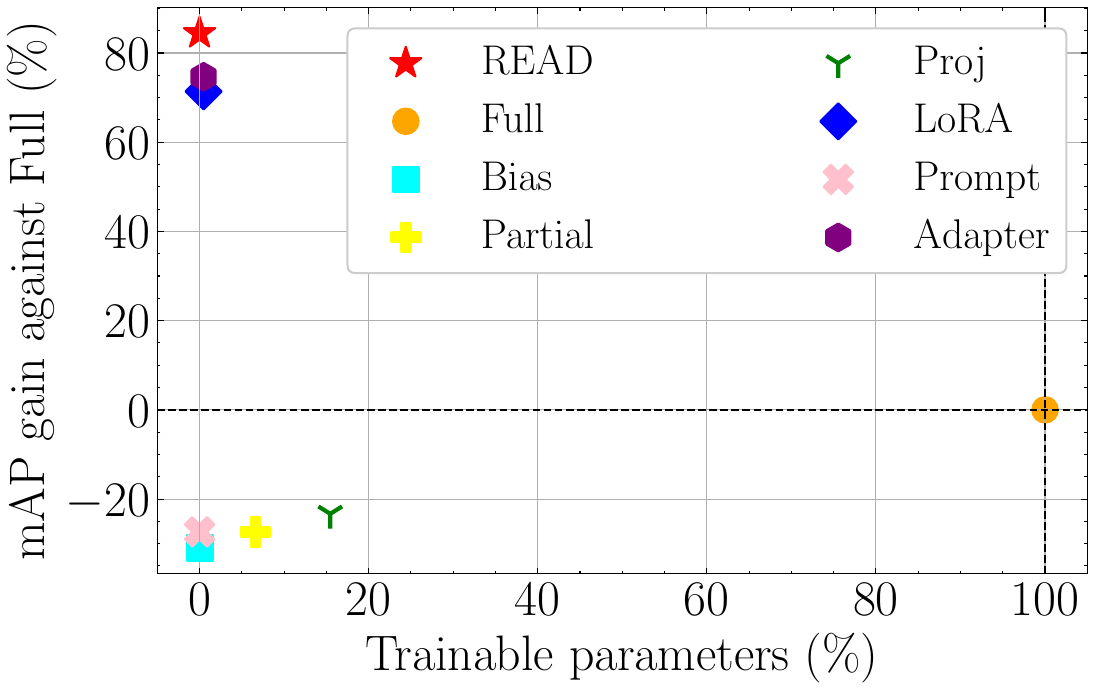}
    \caption{Comparison of our proposed READ method with the full fine-tuning and other parameter-efficient fine-tuning methods. For each method, we denote the mAP gain averaged over the domains of the YouTube Highlights dataset together with the number of trainable parameters.
    % \wxb{The legends seem to be data points in the figure.}
    }
    \label{fig:map_gain_trainable_parameters}
    \vspace{-10pt}
\end{figure}
% To resolve the first issue, we propose \textbf{Re}current \textbf{Ad}atper (READ) to address the capability of modeling fine-grained temporal information. Particularly, after embedding video-language into a sequence of hidden representations, we introduce a novel parameter-efficient bottleneck operation formulated as a procedure of feature dimension reduction, recurrent computation, and feature dimension recovery. We designate the recurrent computation over the sequence length to learn the temporal dependencies among video-language tokens, thus capturing the fine-grained temporal dependencies for enhanced video-language modeling. \zhiyuan{I think we can emphasize this compression and recovery adapter can capture the task-specific temporal feature so that it can be efficiently work in a new low resource scenario. But why we need to mention fine-grained temporal dependencies? what is its precious meaning and why READ can capture fine-grained feature rather than coarse-grained information}
To resolve the first issue, we propose a novel adapter architecture, \textbf{RE}current \textbf{AD}apter (READ), for video-language modeling tasks. The key idea is to incorporate the recurrent modeling ability into the adaptation module to capture the temporal dependency of video frames and language entities \citep{goodfellow2016deep}. As such, we formulate READ as a parameter-efficient bottleneck with a sequence of operations including feature dimension reduction, recurrent modeling, and feature dimension recovery. Since the incorporated recurrent computation works in the low dimension (\textit{e.g.} 4-dimensional), our READ module stands as a lightweight design and can be cheaply integrated throughout the Transformer architecture for enhancing video-language modeling, using only up to 1.20\% trainable parameters.

As for the second issue, we propose \textbf{P}artial \textbf{V}ideo-\textbf{L}anguage \textbf{A}lignment (PVLA), a novel objective to explicitly encourage the alignment between video and language representations, thus capturing invariant aligned information across modalities that are critical for downstream tasks. The key concept is to minimize the Partial Optimal Transport (POT) distance between the distribution over video frame representations and the distribution over textual word representations. The rationale for our partial implementation of optimal transport lies in that video and language do not exhibit complete one-to-one correspondence. Typically, the language does not describe all aspects of the video, and only part of the language sequence is strongly related to part of the video frames, \textit{e.g.} in Figure \ref{fig:task_example} the language input about the \textit{girl’s expression} is only related to the target grounding. As such, utilizing POT for distribution matching is to focus on essential masses that are strongly related between modalities, hence optimizing towards better video-language alignment and gaining more control over video-language information passed into our READ modules.

Based on our novel proposals, we construct the READ framework that can be employed to finetune various pre-trained Transformer architectures such as multimodal transformer (UMT \citep{liu2022umt}, Moment-DETR \citep{lei2021detecting}), and generative vision-guided transformer models (VG-BART \citep{lewis2020bart} and VG-T5 \citep{raffel2020exploring}). %
Through freezing these pre-trained models and fine-tuning only our READ modules with PVLA objective, we outperform standard fine-tuning and other parameter-efficient methods with substantially fewer tunable parameters (Figure \ref{fig:map_gain_trainable_parameters}) for low-resource video-language tasks, including temporal language grounding and video-language summarization. To sum up, our contributions can be summarized as:
\begin{itemize}
    \item We propose \textbf{RE}current \textbf{AD}apter (READ), a novel adapter architecture,
    that better captures  temporal information for modeling video-language tasks.
    \item We propose \textbf{P}artial \textbf{V}ideo-\textbf{L}anguage \textbf{A}lignment (PVLA) objective to encourage the alignment between video and language modalities during the adaptation process.
    \item We validate our READ framework by extensive experiments using multiple low-resource temporal language grounding and video-language summarization datasets,
     where READ outperforms all existing fully or parameter-efficient fine-tuning strategies with only up to 1.20\% parameters tunable.
\end{itemize}

%% file: files/02_related_work.tex
\section{Related Work}
\subsection{Parameter-Efficient Transfer Learning}
% Motivated by the prevalent application of pre-trained Transformer models, 
Recent efforts have sought to propose techniques to reduce the cost of fine-tuning these large-scale models. The techniques can be categorized into three directions. The first one, dubbed as adapter, introduces lightweight modules between Transformer layers that work in low-dimensional space \citep{houlsby2019parameter, pan2022st, chen2022adaptformer, xu2023side}. During fine-tuning, only parameters of the adapters are updated and all of the original Transformer are kept frozen. The second approach, called prompt tuning, appends a sequence of prefix continuous tokens to every input and solely tunes these tokens for adapting to the downstream task \citep{jia2022visual, huang2023vop}. The third approach approximates the weight update with low-rank matrices \citep{hu2021lora}. Only values of the matrices are learned during training to satisfy the parameter-efficiency requirement. 

% Concerning three approaches, the prompt tuning method demands manual labor to select appropriate number of tokens and initialization schemes to avoid training unstability. For video-language downstream tasks that typically involve multiple input modality streams, such demand is labor-intensive and could bring in sub-optimal performance.
\subsection{Video-Language Modeling}
Recent video-language modeling tasks, \emph{e.g.} temporal language grounding (TLG) \citep{liu2022umt, lei2021detecting, nguyen2023demaformer} or video-language summarization (VLS) \citep{yu2021vision, liu2023long} have been dominated by deep Transformer models. Regarding the TLG task, \citep{lei2021detecting} collect a query-based benchmark and construct a Transformer model pre-trained upon automatic speech recognition to tackle not only their benchmark but also other datasets. \citep{liu2022umt} follow with a multimodal Transformer that first considers video and language as separate input streams and unifies them, while preserving the pre-training scheme of \citep{lei2021detecting}. For the VLS task, \citep{yu2021vision} take into account generative pre-trained Transformer to fuse the video and language content then generate the output summary. 
% Although these methods have achieved promising results, they mostly rely on either large-scale training data or huge-parameter adaptation, which hinders their applicability in practice. In this work, we study methods to more efficiently perform adaptation of these models to real-life cases, especially when the training data is of limited scale and computing resource is restricted.

%% file: files/03_methodology.tex
\begin{figure}[t]
    \centering
\includegraphics[width=0.8\linewidth]{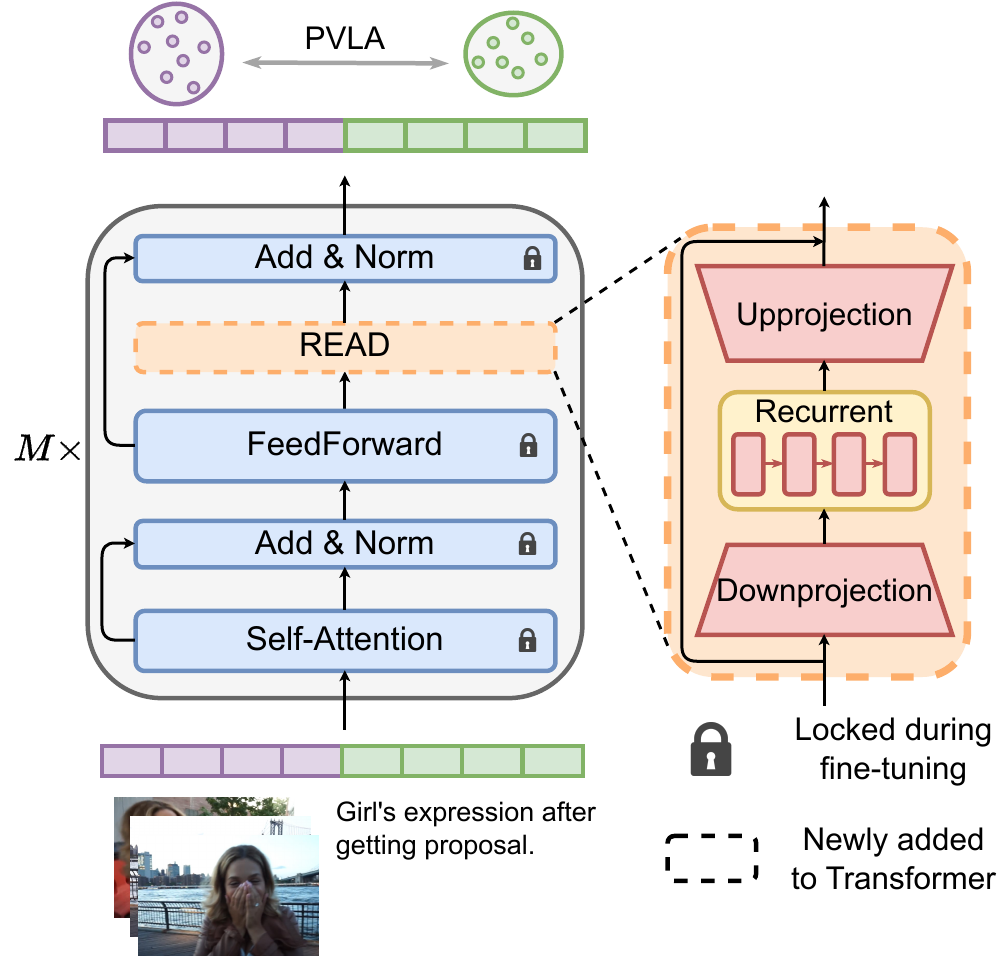}
    \caption{Overall illustration of the proposed recurrent adapter (READ) and partial video-language alignment (PVLA) framework.} 
    \label{fig:overall_illustration}
    \vspace{-20pt}
\end{figure}
\vspace{-5pt}
\section{Methodology}
We present recurrent adapter (READ) to effectively develop the temporal modeling capability and efficiently transfer large pre-trained transformer models for video-language downstream tasks. We also introduce the partial video-language alignment (PVLA) task to optimize the alignment of in-distribution video-language inputs for better supporting video-language adaptation under low-resource settings. Our overall framework is illustrated in Figure \ref{fig:overall_illustration}.

\subsection{Preliminary – Transformer Architecture for Video-Language Modeling}
\label{sect:preliminary}
We concentrate our work upon the Transformer architecture \citep{vaswani2017attention}. The architecture consists of an embedding layer and $M$ consecutive Transformer blocks. As inputs to the Transformer model, we extract $N_{V}$ frames and $N_{L}$ words from the video and language input, respectively. The embedding layer would encode the extracted frames and words into sequences of initial video and language representations $H_{V}^{(0)} = \{\mathbf{h}_{v,i}^{(0)}\}_{i=1}^{N_V}$ and $H_{L}^{(0)} = \{\mathbf{h}_{l,j}^{(0)}\}_{j=1}^{N_L}$, respectively. The transformer then forwards these sequences into consecutive Transformer blocks, each of which is typically composed of a multi-head self-attention (MHSA) layer, a residual connection with normalization (Add \& Norm) layer, a feedforward layer, and another Add \& Norm layer. 

% \wxb{It seems unnecessary to introduce too much about Transformers.}
In MHSA for video-language modeling, the language representations are linearly projected into the query tensor $\mathbf{Q} \in \mathbb{R}^{N_L \times d}$, whilst the video representations into the key $\mathbf{K} \in \mathbb{R}^{N_V \times d}$ and value tensors $\mathbf{V} \in \mathbb{R}^{N_V \times d}$:
\begin{gather*}
\mathbf{Q}^{(m)} = \text{Linear}\left(\mathbf{H}_{L}^{(m)}\right), \; \mathbf{K}^{(m)} = \text{Linear}\left(\mathbf{H}_{V}^{(m)}\right), \\
\mathbf{V}^{(m)} = \text{Linear}\left(\mathbf{H}_{V}^{(m)}\right),
\end{gather*}
where $m$ denotes the index of the current Transformer block and $d$ the hidden dimension. Then, the self-attention computation is conducted upon these vectors as:
\begin{equation}
\begin{split}
\mathbf{X}^{(m)} = \text{Attention}\left(\mathbf{Q}^{(m)}, \mathbf{K}^{(m)}, \mathbf{V}^{(m)}\right) = \\ \text{Softmax}\left(\frac{\mathbf{Q}^{(m)}\cdot(\mathbf{K}^{(m)})^{\top}}{\sqrt{d}}\right) \cdot \mathbf{V}^{(m)}.
\end{split}
\end{equation}
The attention output $\mathbf{X}^{(m)}$ is subsequently sent to an Add \& Norm layer:
\begin{equation}
\mathbf{P}^{(m)} = \text{LN}\left(\mathbf{X}^{(m)} + \mathbf{H}^{(m)}_{L}\right),
\end{equation}
where LN denotes the layer normalization layer. Subsequently, $\mathbf{P}^{(m)}$ is forwarded to a FeedForward block to produce the output representation $\mathbf{O}^{(m)}$, which will be passed to another Add \& Norm layer to create the video-informed language representation for the next transformer block:
\begin{gather}
\mathbf{O}^{(m)} = \text{GeLU}\left(\text{Linear}\left(\mathbf{P}^{(m)}\right)\right), \\ 
\mathbf{H}_{L}^{(m+1)} = \text{LN}\left(\mathbf{P}^{(m)} + \mathbf{O}^{(m)} \right), \; \mathbf{H}_{V}^{(m+1)} = \mathbf{H}_{V}^{(m)}.
\end{gather}
The video-language representation of the last Transformer block $\mathbf{H}_{L}^{(M+1)}$ is finally adopted to perform a specific downstream task.

\subsection{Recurrent Adapter (READ)}
The objective of our READ is to incorporate the temporal modeling capability for the adaptation module. To this end, we construct a recurrent-based bottleneck layer which is composed of a downprojection layer, a recurrent neural network (RNN) layer, and an up-projection layer.

Formally, given the FeedForward output $\mathbf{O}$, our recurrent adapter can be expressed as:
% \vspace{-4pt}
\begin{gather}
\mathbf{\tilde{O}} = \mathbf{O} + \text{GELU}\left(\text{RNN}\left(\mathbf{O} \cdot W_{\text{down}}\right)\right)\cdot W_{\text{up}},
\end{gather}
where $W_{\text{down}} \in \mathbb{R}^{d \times k}, W_{\text{up}} \in \mathbb{R}^{k \times d}$, and $k \ll d$. Subsequently, we combine $\mathbf{P}$ and $\tilde{\mathbf{O}}$ via residual connection to generate the output $\mathbf{H}$:
\begin{equation}
\mathbf{H} = \text{LN}\left(\mathbf{\tilde{O}} + \mathbf{P}\right).
\end{equation}
In addition to RNN, we also experiment with other recurrent architectures in Table \ref{tab:recurrent_ablation_experiments} and observe that the performance is insensitive to the architectural choice. Therefore, for simplicity, we decide to implement the RNN architecture in our READ layer.
% With the RNN operator\zhiyuan{is it the vanilla RNN? why not LSTM or GRE which can avoid the problem in long sequence. video can a very long sequence --> also found this part later. Do we briefly clarify different setting here}, we enable the temporal inference efficiently, because the operator only works in a compressed low-dimensional (\textit{e.g.} 4D) feature space. Empirically, our proposal yields an introduction of tiny additional ($\sim 1\%$) parameters and ($\sim 0.3\%$) computation.

\noindent\textbf{Fine-tuning.} During the fine-tuning stage, we preserve the weights of the pre-trained Transformer model and only optimize our introduced READ layers. In detail, the original model components (blue blocks in Figure \ref{fig:overall_illustration}) are frozen, while the parameters of READ (the yellow block in Figure \ref{fig:overall_illustration}) are updated with respect to the task-specific and the partial video-language alignment losses, which will be delineated in the upcoming section.

\noindent\textbf{Testing.} During testing, we maintain the shared parameters of the pre-trained Transformer model and only load those of our extra READ modules that are fine-tuned in the previous phase. This would keep the storage cost from burgeoning because the number of added parameters is tiny.
% \vspace{-5pt}
\subsection{Partial Video-Language Alignment (PVLA)}
To encourage the control towards the information flow of video frames and language words, we propose to optimize the alignment between the in-distribution video and language representations $H_{V}$ and $H_{L}$ at all Transformer blocks. 

We consider video and language as two discrete distributions $\boldsymbol{\mu}$ and $\boldsymbol{\nu}$, whose $H_V$ and $H_L$ are their supports, respectively.  We formulate this setting as $\boldsymbol{\mu} = \sum\limits_{i=1}^{N_V} \mathbf{a}_{i} \delta_{\mathbf{h}_{v,i}}$ and $\boldsymbol{\nu} = \sum\limits_{j=1}^{N_L} \mathbf{b}_{j} \delta_{\mathbf{h}_{l,j}}$, with $\delta_{\mathbf{h}_{l,j}}$ and $\delta_{\mathbf{h}_{l,j}}$ being the Dirac functions respectively centered upon $\mathbf{h}_{v,i}$ and $\mathbf{h}_{l,j}$. The weight vector of the supports is $\mathbf{a} = \frac{\mathbf{1}_{N_V}}{N_V}$, and $\mathbf{b} = \frac{\mathbf{1}_{N_L}}{N_L}$.

Based upon the above setting, we propose the partial video-language alignment (PVLA) task, which is to minimize the following $\mathcal{L}_{\text{PVLA}}$ loss equal to the partial optimal transport (POT) distance $\mathcal{D}_{\text{POT}}$ between $\boldsymbol{\mu}$ and $\boldsymbol{\nu}$ as:
\begin{gather}
\mathcal{L}_{\text{PVLA}} = \mathcal{D}_{\text{POT}}(\boldsymbol{\mu}, \boldsymbol{\nu}) = \min_{\mathbf{T} \in \Pi(\mathbf{a}, \mathbf{b})} \sum\limits_{i=1}^{N_V} \sum\limits_{j=1}^{N_L} \mathbf{T}_{i,j} \cdot c\left(\mathbf{h}_{v,i}, \mathbf{h}_{l,j}\right), \\
\begin{split}
\text{s.t} \quad \Pi(\mathbf{a}, \mathbf{b}) = \{\mathbf{T} \in \mathbb{R}^{N_V \times N_L}_{+} \mid \mathbf{T}\mathbf{1}_{N_L} \leq \mathbf{a},  \mathbf{T}^{\top}\mathbf{1}_{N_V} \leq \mathbf{b}, \\\mathbf{1}_{N_V}^{\top} \cdot \mathbf{T} \cdot \mathbf{1}_{N_L} = s, \quad 0 \leq s \leq \min(N_L, N_V)\}.
\label{eq:pvla_formulation}
\end{split}
\end{gather}
Because the exact minimization over the transport plan $\mathbf{T}$ is intractable, we adopt the Sinkhorn-based algorithm to compute $\mathbf{T}$. We explicate our algorithm to calculate the partial video-language alignment loss in Algorithm \ref{alg:pvla_loss}.

Our PVLA formulation is flexible where it allows only $s$ samples from one distribution to be transported to the other, and enables the algorithm to decide the value of $s$, in case the input language only corresponds to certain video aspects (or vice versa).

\noindent\textbf{Training Strategy.} For training, we jointly optimize the video-language task-specific loss and our PVLA loss. It is worth noting that we only update our introduced READ layers while keeping the remaining components frozen.

%% file: files/04_experiments.tex
\section{Experiments}
\setlength{\textfloatsep}{10pt}
\begin{algorithm}[t]
\caption{Computing the PVLA loss}
\label{alg:pvla_loss}
\begin{algorithmic}
\Require{$\mathbf{C} = \{\mathbf{C}_{i,j} = c\left(\mathbf{h}_{v,i}, \mathbf{h}_{l,j}\right) \mid 1 \leq i \leq N_V, 1 \leq j \leq N_L\} \in \mathbb{R}^{N_V \times N_L}$,\; temperature $\tau,\; \mathbf{a} \in \mathbb{R}^{N_V},\; \mathbf{b} \in \mathbb{R}^{N_L},\; s,\; N_{\text{iter}}$} \\
$\mathcal{L}_{\text{PVLA}} = \infty$ 
\For{$s=1$ to $\min(N_L, N_V)$}
    \State $\mathbf{T} = \text{exp}\left(-\frac{\mathbf{C}}{\tau}\right)$ 
    \State $\mathbf{T} = \frac{s}{\left(\mathbf{1}_{N_V}\right)^{\top} \cdot \mathbf{T} \cdot \mathbf{1}_{N_L}} \mathbf{T}$
    
    \For{$i=1$ to $N_{\text{iter}}$} 
        \State $\mathbf{p}_{a} = \min \left(\frac{\mathbf{a}}{\mathbf{T}\mathbf{1}_{N_L}}, \mathbf{1}_{N_V}\right)$, $\mathbf{T}_{a} = \text{diag}\left(\mathbf{p}_{a}\right) \cdot \mathbf{T}$
        \State $\mathbf{p}_{b} = \min \left(\frac{\mathbf{b}}{\mathbf{T}_{a}^{\top}\mathbf{1}_{N_V}}, \mathbf{1}_{N_L}\right)$, $\mathbf{T}_{b} = \text{diag}\left(\mathbf{p}_{b}\right) \cdot \mathbf{T}_{a }$
        \State $\mathbf{T} = \frac{s}{\left(\mathbf{1}_{N_V}\right)^{\top} \cdot \mathbf{T} \cdot \mathbf{1}_{N_L}} \mathbf{T}_{b}$
    \EndFor 
    \State $\mathcal{L}_{\text{PVLA}} = \min\left(\mathcal{L}_{\text{PVLA}}, \;\sum\limits_{i=1}^{N_V} \sum\limits_{j=1}^{N_L} \mathbf{T}_{i,j} \mathbf{C}_{i,j}\right)$ 
    \EndFor \\
\Return $\mathcal{L}_{\text{PVLA}}$
\end{algorithmic}
\end{algorithm}

{\renewcommand{\arraystretch}{1.2}
\begin{table*}[h!]
\centering
% \resizebox{0.5\linewidth}{!}{
\fontsize{9pt}{9pt}\selectfont
\begin{tabular}{l|c|ccccccc}
\hline
\multicolumn{1}{c|}{\textbf{Method}} & \multicolumn{1}{c|}{\centering\textbf{\#params (M)}} & \multicolumn{1}{c}{\textbf{Dog}} & \multicolumn{1}{c}{\textbf{Gym}} & \multicolumn{1}{c}{\textbf{Par.}} & \multicolumn{1}{c}{\textbf{Ska.}} & \multicolumn{1}{c}{\textbf{Ski.}} & \multicolumn{1}{c}{\textbf{Sur.}} & \multicolumn{1}{c}{\textbf{Avg.}} \\ \hline
Full                       &       283.97 (100\%)                                   &      65.90\sig                   &      75.20\sig                   &        82.20\sig                  &         71.80\sig                 &         72.30\sig                 &       81.15\sig                   &              74.76\sig            \\
Bias                       &      0.51 (0.18\%)                                    &       46.23\sig                  &     61.19\sig                    &          56.73\sig                &         31.36\sig                 &          61.14\sig                &      49.77\sig                    &           51.07\sig               \\
Partial                    &       38.75 (13.65\%)                                   &       48.28\sig                  &       63.26\sig                  &           59.71\sig               &           32.66\sig               &           64.58\sig               &        56.22\sig                  &        54.12\sig                  \\
Proj                       &        5e-4 (1.76e-4\%)                                  &       57.05\sig                  &    65.70\sig                     &          63.03\sig                &        71.83\sig                  &            65.45\sig              &           79.71\sig               &         67.13\sig                 \\
LoRA                       &            13.12 (4.62\%)                             &               60.97\sig          &          67.68\sig               &          72.53\sig                &          66.62\sig                &              71.24\sig            &    79.15\sig                      &            69.70\sig              \\
Prompt                       &      0.02 (0.01\%)                                    &               48.28\sig          &               63.26\sig        &       59.71\sig                   &           35.67\sig               &       35.67\sig                   &        64.61\sig                  &           46.87\sig               \\
Adapter                    &       13.11 (4.62\%)                                   &         62.89\sig                &       67.09\sig                  &       74.56\sig                   &           62.56\sig               &        68.10\sig                  &                    78.73\sig      &          68.98\sig                \\ \hline
READ                     & 0.16 (0.06\%)                   & \textbf{67.65}                   & \textbf{78.05}                   & \textbf{83.25}                    & \textbf{72.40}                    & \textbf{72.98}                    & \textbf{82.36}                    & \textbf{76.12}  \\ \hline
    \end{tabular}
    % }
\caption{TLG results on the YouTube Highlights dataset. We report the mean average precision (mAP) and the number of trainable parameters (\#params). \sig means the gain of READ is statistically significant at the 0.05 level.}
\label{tab:youtube_highlights_results}
\end{table*}}
{\renewcommand{\arraystretch}{1.2}
\begin{table*}[h!]
\centering
% \resizebox{0.65\linewidth}{!}{
\fontsize{9pt}{9pt}\selectfont
\begin{tabular}{p{0.11\linewidth}|>{\centering\arraybackslash}p{0.12\linewidth}|>{\centering\arraybackslash}p{0.04\linewidth}>{\centering\arraybackslash}p{0.04\linewidth}>{\centering\arraybackslash}p{0.04\linewidth}>{\centering\arraybackslash}p{0.04\linewidth}>{\centering\arraybackslash}p{0.04\linewidth}>{\centering\arraybackslash}p{0.04\linewidth}>{\centering\arraybackslash}p{0.04\linewidth}>{\centering\arraybackslash}p{0.04\linewidth}>{\centering\arraybackslash}p{0.04\linewidth}>{\centering\arraybackslash}p{0.04\linewidth}>{\centering\arraybackslash}p{0.04\linewidth}}
\hline
\textbf{Method} & \textbf{\#params (M)} & \textbf{VT} & \textbf{VU} & \textbf{GA} & \textbf{MS} & \textbf{PK} & \textbf{PR} & \textbf{FM} & \textbf{BK} & \textbf{BT} & \textbf{DS} & \textbf{Avg.} \\ \hline
Full            &         285.28 (100\%)                        &     84.17\sig                            &           81.50\sig                      &           88.20\sig                      &             71.54\sig                    &         81.40\sig                        &              84.31\sig                   &         72.30\sig                        &             76.53\sig                    &           78.86\sig                      &             77.70\sig  &   79.65\sig                      \\
Bias            &             0.25 (0.09\%)                    &                         38.08\sig        &                   69.62\sig              &         60.87\sig                        &             31.25\sig
&     68.84\sig                            &               51.71\sig                  &        50.72\sig                         &          65.38\sig                       &         54.42\sig                        &            59.05\sig & 54.99\sig                       \\
Partial         &             38.75 (13.58\%)                    &                   57.27\sig              &            62.57\sig                     &                      58.08\sig           &            52.35\sig                     &         61.58\sig                        &               63.94\sig                  &        50.82\sig                         &            62.36\sig                     &        58.05\sig                         &         47.79\sig & 57.48\sig                          \\
Proj            &            5e-4 (1.75e-4\%)                     &       57.65\sig                          &        65.80\sig                         &      64.40\sig                           &          55.57\sig                       &           64.67\sig                      &         67.07\sig                        &              59.08\sig                   &            74.70\sig                     &        63.29\sig                         &                  49.48\sig & 62.17\sig                 \\
LoRA            &      13.28 (4.66\%)                           &        77.87\sig                         &        77.01\sig                         &        77.82\sig                         &           66.38\sig                      &       80.21\sig                          &           82.23\sig                      &         66.89\sig                        &                  72.31\sig               &       69.58\sig                          &           72.09\sig & 74.24\sig                        \\
Prompt            &     0.02 (0.007\%)                            &                    61.67\sig             &         71.98\sig                        &          64.07\sig                       &              35.54\sig                   &                      72.74\sig           &              48.70\sig                   &      52.97\sig              & 67.59\sig             &         57.28\sig                        &           38.60\sig                      &          57.11\sig                         \\
Adapter         &            13.29 (4.66\%)                     &      78.46\sig                           &       76.38\sig                          &           77.36\sig                      &            67.12\sig                     &           80.33\sig                      &             82.51\sig                    &            67.77\sig                     &           71.71\sig                      &          69.58\sig                       &           71.24\sig & 74.25\sig                        \\ \hline 
READ     &  0.14 (0.05\%)      & \textbf{88.30}                           & \textbf{85.15}                           & \textbf{89.76}                           & \textbf{75.80}                           & \textbf{86.69}                           & \textbf{86.62}                           & \textbf{74.99}                           & \textbf{82.38}                           & \textbf{84.65}                           & \textbf{79.60}                           & \textbf{83.39}   \\                  \hline 
\end{tabular}
% }
\caption{TLG results on the TVSum dataset. We report the mean average precision (mAP) and the number of trainable parameters (\#params). \sig means the gain of READ is statistically significant at the 0.05 level.}
\label{tab:tvsum_results}
\vspace{-5pt}
\end{table*}}

{\renewcommand{\arraystretch}{1.2}
\begin{table}[h!]
\centering
% \resizebox{0.5\linewidth}{!}{
\fontsize{9pt}{9pt}\selectfont
\begin{tabular}{l|c|c}
\hline
\textbf{Method} & \textbf{\#params (M)} & \textbf{mAP} \\ \hline
Full            &           15.88 (100\%)                    &      36.14\sig        \\
Bias            &          0.06 (0.38\%)                     &       24.89\sig       \\
Partial         &         1.05 (6.61\%)                      &     26.37\sig         \\
Proj            &          7.31 (46.03\%)                     &   32.71\sig           \\
LoRA            &      0.19 (1.20\%)                       &     33.96\sig         \\
Prompt          &      0.04 (0.25\%)                         &    25.86\sig          \\
Adapter         &         0.20 (1.26\%)                      &       33.61\sig       \\ \hline
READ       &          0.19 (1.20\%)                     &     \textbf{36.74}        \\ \hline
\end{tabular}
% }
\caption{TLG results on the QVHighlights dataset. We report the mean average precision (mAP) and the number of trainable parameters (\#params). \sig means the gain of READ is statistically significant at the 0.05 level.}
\label{tab:qvhighlights_resultss}
% \vspace{-10pt}
\end{table}}

We conduct extensive experiments to evaluate the effectiveness of our READ framework. We first describe the experimental settings, covering the downstream tasks, evaluation metrics, pre-trained backbones, baseline approaches, and implementation details. We then present the numerical results of our method with baseline models, then provide ablation study and thorough analysis to explore various configurations. Eventually, we perform qualitative assessments to further elucidate the behavior of our framework.
\subsection{Experimental Settings}
\label{subsect:experimental_settings}
\noindent\textbf{Downstream tasks.} We assess the effectiveness on the temporal language grounding and video-language summarization tasks. The corresponding datasets to each task are presented as follows:
\begin{itemize}
    \item \emph{Temporal Language Grounding (TLG):} The TLG’s task is to localize temporal boundaries of the video frames that semantically relate to the language query. The evaluation is performed upon three datasets, \emph{i.e.} YouTube Highlights \citep{sun2014ranking}, TVSum \citep{song2015tvsum}, and QVHighlights \citep{lei2021detecting}. YouTube Highlights consists of 40 video-language training inputs for each of the 6 domains. TVSum comprises 10 domains, each of which possesses 5 video-language training inputs. The QVHighlights benchmark includes 7,218 language-annotated video segments for training, 1,550 for development, and 1,542 for testing. 
    Following previous work on low-resource experiments \citep{boulanger2022generating}, we keep our training size at 700 samples, which is less than 10\% of the full data for the QVHighlights dataset, while preserving the original splits on the TVSum and YouTube Highlights datasets.
    % Because these dataset sizes already satisfy the low-resource level in terms of today’s deep learning era, we preserve their original splits for our experiments.
    \item \emph{Video-Language Summarization (VLG):} Given a video-language input, the VLS’s target is to generate a summary which takes into account both video and language content \citep{yu2021vision}. We consider the How2 dataset \citep{sanabria2018how2}, from which we randomly draw 2,000 out of 73,993 samples for training, \textit{i.e.} less than 3\% of the full data, to simulate the low-resource settings, while maintaining 2,520 samples for validation, and 2,127 samples for testing.
\end{itemize}
\textbf{Evaluation metrics.} For the TLG task, we follow previous works \citep{lei2021detecting, liu2022umt} to use the mean average precision (mAP) metric. Regarding VLS, we utilize the ROUGE score, which is a popular metric for summarization \citep{zhang2020pegasus, yu2021vision}.

{\renewcommand{\arraystretch}{1.2}
\begin{table}[h!]
\centering
% \resizebox{0.65\linewidth}{!}{
\fontsize{9pt}{9pt}\selectfont
\begin{tabular}{l|>{\centering\arraybackslash}p{0.24\linewidth}|>{\centering\arraybackslash}p{0.1\linewidth}>{\centering\arraybackslash}p{0.1\linewidth}>{\centering\arraybackslash}p{0.1\linewidth}}
\hline
\textbf{Method}    & \multicolumn{1}{p{0.25\linewidth}|}{\centering\textbf{\#params (M)}}  & \textbf{R1} & \textbf{R2} & \textbf{RL} \\ \hline
Full      &          249.67 (100\%)                &  35.72\sig  &  11.88\sig   &   30.00\sig \\
Bias      &              0.20 (0.08\%)            &  30.51\sig   &  8.20\sig  &  23.00\sig  \\
Partial   &            16.54 (6.62\%)              &  31.55\sig  &  8.63\sig  &  13.65\sig  \\
Proj      &        38.60 (15.46\%)                  &  32.76\sig   &  9.11\sig   &  30.01\sig  \\
LoRA      &          1.19 (0.48\%)                &  40.04\sig  &  20.36\sig  & 35.98\sig   \\
Prompt       &       0.05 (0.02\%)                   &  31.55\sig  &  8.63\sig  &  14.90\sig  \\
Adapter   &           1.20 (0.48\%) &  41.52\sig   &  20.75\sig  &  36.88\sig  \\ \hline
READ &            1.17 (0.47\%)            &  \textbf{44.01} &  \textbf{21.91}  &  \textbf{37.91} \\ \hline
\end{tabular}
% }
\caption{VLS results on the How2 dataset with the VG-BART model. We report the ROUGE-1, ROUGE-2, and ROUGE-L scores, with the number of trainable parameters (\#params). \sig means the gain of READ is statistically significant at the 0.05 level.}
\label{tab:how2_bart_results}
\end{table}}

{\renewcommand{\arraystretch}{1.2}
\begin{table}[h!]
\centering
% \resizebox{0.6\linewidth}{!}{
\fontsize{9pt}{9pt}\selectfont
\begin{tabular}{l|c|ccc}
\hline
\textbf{Method}    & \multicolumn{1}{p{0.25\linewidth}|}{\centering\textbf{\#params (M)}}  & \textbf{R1} & \textbf{R2} & \textbf{RL} \\ \hline
Full      &          333.16 (100\%)                &  32.37\sig  & 8.07\sig   &  26.53\sig   \\
Bias      &           0.07 (0.02\%)               &  27.03\sig  &  4.53\sig  & 19.51\sig   \\
Partial   &           16.52 (4.96\%)               &  27.83\sig   &  4.92\sig  &  10.49\sig  \\
Proj      &           24.67 (7.40\%)               &  29.16\sig  &  5.50\sig  &  26.76\sig  \\
LoRA      &        1.93 (0.58\%)                  &  36.63\sig  &  16.72\sig  &  32.77\sig  \\
Prompt       &      0.18 (0.05\%)                &  28.33\sig   &  5.12\sig  &  11.75\sig  \\
Adapter   &           1.95 (0.59\%)               &  37.83\sig  &  17.45\sig  &  33.87\sig   \\ \hline
READ &           1.91 (0.57\%)              &  \textbf{40.12} &  \textbf{18.71}  &  \textbf{34.42} \\ \hline
\end{tabular}
% }
\caption{VLS results on the How2 dataset with the VG-T5 model. We report the ROUGE-1, ROUGE-2, and ROUGE-L scores, with the number of trainable parameters (\#params). \sig means the gain of READ is statistically significant at the 0.05 level.}
\label{tab:how2_t5_results}
\end{table}}

{\renewcommand{\arraystretch}{1.2}
\begin{table}[h!]
\centering
% \resizebox{0.6\linewidth}{!}{
\fontsize{9pt}{9pt}\selectfont
\begin{tabular}{l|cc}
\hline
\textbf{Method} & \textbf{mAP - YouTube Highlights} & \textbf{R2 - How2} \\ \hline
No VLA          &              73.80                     &      18.22              \\
VLA             &               74.41                    &         20.01           \\ 
\rowcolor{Gray}
% \rowcolor{Gray}
PVLA            & 76.12                             & 21.91     \\ \hline     
\end{tabular}
% }
\caption{Partial video-language alignment (PVLA) ablation experiments on YouTube Highlights and How2. We \colorbox{Gray}{color} the settings we implement for our READ method.}
\label{tab:pvla_ablation_experiments}
\end{table}}

{\renewcommand{\arraystretch}{1.2}
\begin{table*}[h!]
\centering
% \resizebox{\linewidth}{!}{
\fontsize{9pt}{9pt}\selectfont
\begin{tabular}{>{\centering\arraybackslash}p{0.3\linewidth}|p{0.3\linewidth}cc}
\hline
\textbf{Video}    & \centering\textbf{Language query}                                            & \textbf{POT distance} & \textbf{AP} \\ \hline
\multirow{2}{*}{\includegraphics[width=0.23\linewidth]{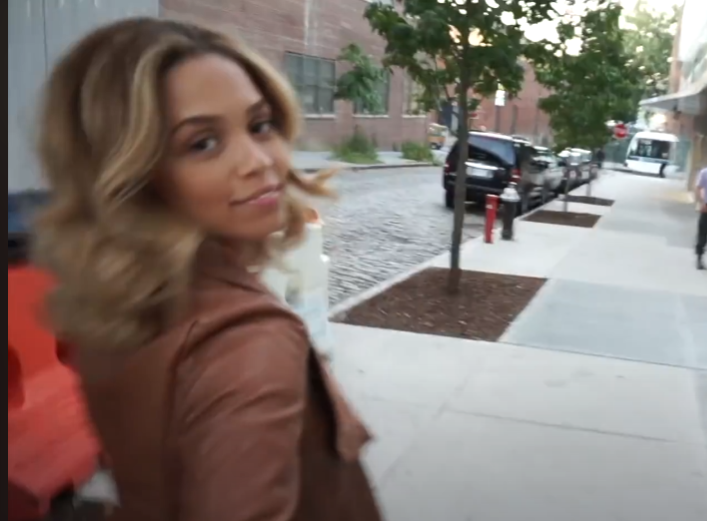} \includegraphics[width=0.23\linewidth]{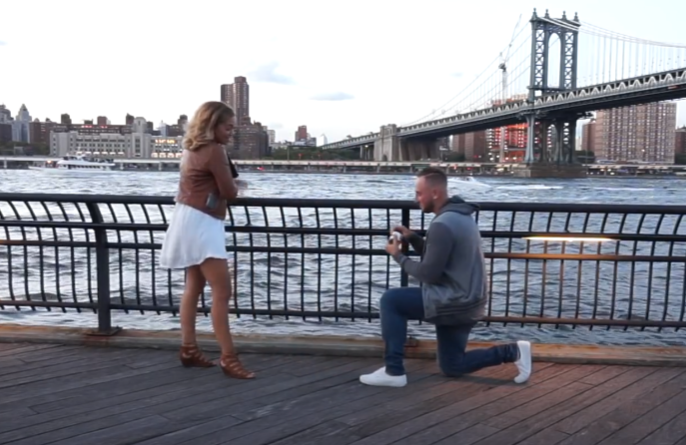}
\includegraphics[width=0.23\linewidth]{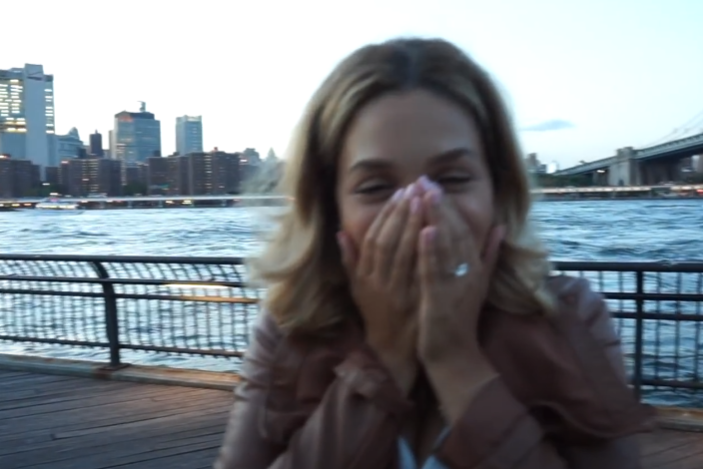} \includegraphics[width=0.23\linewidth]{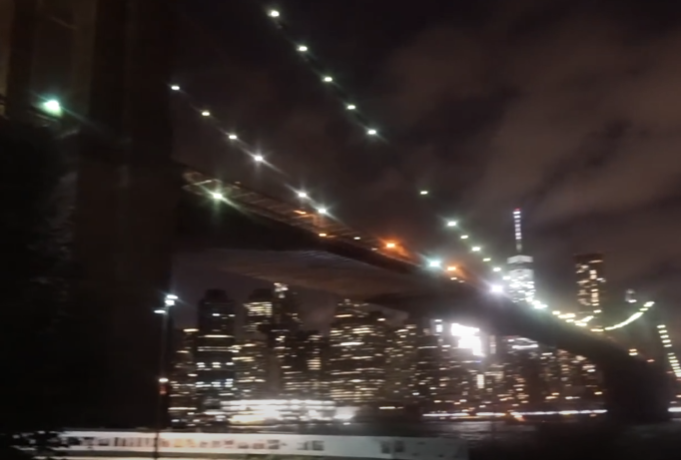}} &  Girl's expression after getting proposal & 27.17                 & 90.28       \\ \cline{2-4}
\multirow{2}{*}{}                  & A boy showing his arm after being stung at the beach                & 56.38                 & 30.00       \\ \hline
\end{tabular}
% }
\caption{Case study on the temporal language grounding benchmark. We extract the POT distance between video and language of two inputs with different language queries and measure the respective AP performance change.}
\label{tab:case_study}
\end{table*}}

{\renewcommand{\arraystretch}{1.2}
\begin{table*}[h!]
\centering
% \resizebox{0.8\linewidth}{!}{
\fontsize{9pt}{9pt}\selectfont
\begin{tabular}{>{\centering\arraybackslash}p{0.2\linewidth}cccc}
\hline
\textbf{Recurrent architecture} & \textbf{\#params - UMT (M)} & \textbf{mAP - YouTube Highlights} & \textbf{\#params - VG-BART (M)} & \textbf{R2 - How2} \\ \hline
\centering GRU                      & 0.16                                      &                   76.07                                    & 1.17                                       &               21.85                         \\
\centering LSTM                     & 0.16                                     &                          76.08                             & 1.17                                       &         21.87                               \\
\rowcolor{Gray}
\centering RNN                      & 0.16                                      & 76.12                                                 & 1.17                                       & 21.91              \\ \hline
\end{tabular}
% }
\caption{Recurrent architecture ablation experiment on YouTube Highlights and How2. We \colorbox{Gray}{color} the settings we implement for our READ method.}
\label{tab:recurrent_ablation_experiments}
\end{table*}}

{\renewcommand{\arraystretch}{1.2}
\begin{table}[h!]
\centering
% \resizebox{0.8\linewidth}{!}{
\fontsize{9pt}{9pt}\selectfont
\begin{tabular}{c|>{\centering\arraybackslash}p{0.3\linewidth}>{\centering\arraybackslash}p{0.2\linewidth}}
\hline
\textbf{Distance method} & \textbf{mAP - YouTube Highlights} & \textbf{R2 - How2} \\ \hline
AvgPool - Cosine         &           71.65                        &           20.32        \\
MaxPool - Cosine         &       74.37                             &      21.36              \\
AvgPool - L2             &             72.11                      &           20.73         \\
MaxPool - L2             &              74.39                     &      21.02              \\
\rowcolor{Gray}
Partial OT                      & 76.12                             & 21.91   \\ \hline          
\end{tabular}
% }
\caption{Distance method ablation experiments on YouTube Highlights and How2. We \colorbox{Gray}{color} the settings we implement for our READ method.}
\label{tab:distance_method_ablation_experiments}
\end{table}}

\noindent\textbf{Pre-trained backbones.} We adopt the Transformer encoder-decoder architecture \citep{vaswani2017attention} pre-trained with both supervised and self-supervised objectives. Specifically, for TLG, we use the unified multimodal transformer (UMT) \citep{liu2022umt} and Moment-DETR \citep{lei2021detecting} models pre-trained upon the automatic speech recognition task. For VLS, we carry out the parameter-efficient adaptation on the generative vision-guided BART (VG-BART) and T5 (VG-T5) \citep{yu2021vision} pre-trained upon reconstruction and masked language modeling tasks \citep{raffel2020exploring}. 

\noindent\textbf{Baseline methods.} We compare our method with a comprehensive list of baseline approaches for efficient video-language transfer learning:
\begin{itemize}
    \item \emph{Full}: update all parameters of the pre-trained backbone.
    \item \emph{Partial}: only update the last layers of the encoder and decoder in the Transformer model.
    \item \emph{Bias} \citep{zaken2022bitfit}: only fine-tune the bias terms in the Transformer backbone.
    \item \emph{Proj}: fine-tune only the last linear projection layer in the Transformer. 
    \item \emph{LoRA} \citep{hu2021lora}: solely fine-tune the decomposition matrices introduced to the linear weights of the Transformer model.
    \item \emph{Prompt} \citep{jia2022visual}: Append a sequence of learnable prompt tokens to both video and language inputs and only fine-tune the appended sequence.
    \item \emph{Adapter} \citep{houlsby2019parameter}: update only the adaptation modules consisting of downprojection and up-projection layers inserted into the Transformer model.
\end{itemize}

\noindent\textbf{Implementation details.} For the TLG task, we use the SlowFast \citep{feichtenhofer2019slowfast} and video encoder of CLIP \citep{radford2021learning} to extract features every 2 seconds. For the VLS task, we use a 3D ResNeXt-101 model to extract a 2048-dimensional embedding for every 16 non-overlapping frames. Similar to previous works \citep{houlsby2019parameter, chen2022adaptformer}, to support training stability, we initialize the weights of the down-projection layer $W_{\text{down}}$ with the Kaiming normal \citep{he2015delving} method, whereas those of the up-projection $W_{\text{up}}$, recurrent layer RNN, and biases of our READ layers are configured with zero initialization. In our PVLA framework, we implement the cost distance $c\left(\mathbf{h}_{v,i}, \mathbf{h}_{l,j}\right)$ as the cosine distance $c\left(\mathbf{h}_{v,i}, \mathbf{h}_{l,j}\right) = 1 - \frac{\mathbf{h}_{v,i} \cdot \mathbf{h}_{l,j}}{||\mathbf{h}_{v,i}||_{2} \cdot ||\mathbf{h}_{l,j}||_{2}}$, and set the maximum number of iterations $N_{\text{iter}}$ to 1,000 and the temperature $\tau$ to 0.05. We fine-tune all models leveraging the AdamW optimizer on 4 NVIDIA Tesla V100 GPUs and report average results of 5 runs. Specific details about the epoch, batch size, learning rate, and the number of Transformer blocks for each task can be found in the Appendix.
\subsection{Main Results}

% \zhiyuan{can we add the ablation study to further enhance our second contribution(the effectiveness of PVLA). only report the READ performance. --> I found this part later. Do we consider add a short summarization about experiments results and analysis here? }

% \zhiyuan{Also, since no any model in baselines can occupy the second-best model position stably, it is better to highlight(like using underline) to clearly illustrate the SOTA performances of baselines and our improvements}
For main comparison of our READ with baseline methods, we denote the results of YouTube Highlights in Table \ref{tab:youtube_highlights_results}, TVSum in Table \ref{tab:tvsum_results}, QVHighlights in Table \ref{tab:qvhighlights_resultss}, and How2 in Tables \ref{tab:how2_bart_results} and \ref{tab:how2_t5_results}.

\noindent\textbf{Temporal language grounding (TLG).} For the YouTube Highlights dataset, our READ framework substantially outperforms the Full fine-tuning approach (\emph{e.g.} 1.36\% on average, 1.75\% in the Dog domain, and 1.21\% in the Surfing domain), while updating far less parameters (0.16M vs. 283.97M). We significantly surpass all other efficient fine-tuning methods as well, \textit{e.g.} with an improvement of 10.96\% over the Adapter in the Gym category, or 5.78\% over LoRA in the Skating category.

For the TVSum dataset, we observe that our method enhances the Full fine-tuning direction with only 0.14M versus 285.28M tunable parameters. For instance, we obtain an increase of 4.26\% in the MS subset and 3.74\% on average. Compared with the best parameter-efficient approach, \emph{i.e.} the Adapter, we achieve a gain of 15.07\% in BT, 10.67\% in BK, and 8.77\% in the VU domain. 

Our improvement also generalizes across different pre-trained backbone. On the QVHighlights dataset, in which we work with the Moment-DETR architecture, we accomplish a gain of 0.6\% over the standard fine-tuning method, while our tunable parameters are only 0.19M versus its 15.88M. We also surpass the efficient approach LoRA with an enhancement of 2.78\% in mAP.

These results demonstrate that our READ framework can efficiently model video-language inputs to polish the low-resource temporal language grounding performance of various pre-trained Transformer models.

\noindent\textbf{Video-language summarization (VLS).} Analogous to the TLG experiment, on the VG-BART backbone, we improve upon the full fine-tuning approach with 8.29 points of ROUGE-1, 10.03 points of ROUGE-2, and 7.91 points of ROUGE-L. Importantly, we only update 1.17M parameters, which account for 0.47\% total parameters of the overall model. On the VG-T5 backbone, we exceed the full approach by 7.75 points in ROUGE-1, 10.64 points in ROUGE-2, and 7.89 points in ROUGE-L, whilst keeping 99.43\% parameters frozen. 

In addition, our framework substantially outperforms other fine-tuning strategies, \emph{e.g.} LoRA with 3.97 points in ROUGE-1, 1.55 points in ROUGE-2, and 1.93 points in ROUGE-L on the VG-BART architecture, along with 3.49 points in ROUGE-1, 1.99 points in ROUGE-2, and 1.65 points in ROUGE-L on the VG-T5 one. 

These results substantiate that our method is applicable to diverse benchmarks and model architectures, particularly not only multimodal Transformers for temporal language grounding but also generative Transformers for video-language summarization. We hypothesize that our advantages are due to the recurrent adapter’s ability to model temporal information and the PVLA task to align video and language signals to maintain more essential information during the efficient fine-tuning stage.
\subsection{Ablation Studies}
We ablate our READ framework to discover what factors result in the demonstrated efficiency and observe several intriguing properties. Our ablation studies are all conducted on the YouTube Highlights and How2 test set. 

\noindent\textbf{Effects of video-language alignment.} We evaluate our framework without the assistance of the PVLA task and with the one of the VLA variant that requires all masses of one distribution to be transferred (we set $s = \min\left(N_V, N_L\right)$ in formulation (\ref{eq:pvla_formulation})). As shown in Table \ref{tab:pvla_ablation_experiments}, the performance drops dramatically when we remove the PVLA task from the fine-tuning procedure. We conjecture that the model has become deficient in managing the information injected into the low-dimensional space of the READ layers, thus passing detrimental noise to the downstream task. Moreover, the VLA variant brings slight performance decrease, which could be due to the VLA’s restrictive nature of transporting all masses from the language distribution to the video one or vice versa.

% \noindent\textbf{Training parameter efficiency.} The bottleneck dimension $k$ of our READ module controls how many parameters are introduced to the pre-trained backbones. Higher dimensions provide more parameters with higher risk of overfitting. In this section, we investigate such effect on the recurrent adapter and denote the results in Table \ref{tab:bottleneck_dimension_ablation_experiments}. As can be seen, the performance consistently improves when the dimension increases up to 4 and saturates after this point. We note that our READ framework can accomplish a decent performance even when the dimension reduces to one, \emph{i.e.} we achieve a ROUGE-2 score of 18.57, verifying the parameter-efficiency of our proposed method. 

\noindent\textbf{Effects of the recurrent architecture.} In addition to RNN, there exist various recurrent architectures in the literature, particularly the gated recurrent unit (GRU) \citep{cho2014learning} and long short-term memory (LSTM) \citep{hochreiter1997long}. We experiment with different recurrent choices and explicate the results in Table \ref{tab:recurrent_ablation_experiments}. As can be observed, the performance is insensitive to the choice of recurrent design. Therefore, we select the simplest option, \emph{i.e.} recurrent neural network (RNN) for our READ layers.

% \noindent\textbf{READ position.} By default, we insert our READ layer to every Transformer block. In this ablation, we demonstrate the impact of using fewer recurrent adapters. Table \ref{tab:position_ablation_experiments} shows that whereas more injections of READ are more effective, the placement of READ at deeper blocks brings higher performance boost than at shallower ones. This discovery can become valuable if we have tight resource and providing a READ layer for every block is expensive. Furthermore, we denote the performance when adapting READ to various positions within a Transformer block. Table \ref{tab:position_within_transformer_block_experiments} reveals that after feedforward is the optimal position for our READ layer, whose observation aligns with previous works \citep{bapna2019simple, pfeiffer2020adapterfusion}.

\noindent\textbf{Distance methods.} We further ablate on the distance metrics to estimate the distance between video and language distributions. Technically, we perform the average- and max-pooling of the video and language representations. Then, we consider the cosine distance or the L2 distance of the two pooled vectors as the video-language distance. Results in Table \ref{tab:distance_method_ablation_experiments} substantiate the superiority of our POT distance for the PVLA objective. Such success illustrates the POT-based PVLA’s advantage of modeling the relationship nature between video and language representations
\subsection{Qualitative Assessment}
\noindent\textbf{Case study.} We display a TLG example on the YouTube Highlights dataset, along with the POT distance estimated by our PVLA framework and the AP score in Table \ref{tab:case_study}. We observe that when the language query semantically corresponds to a moment in the video, \textit{i.e. a girl expression after she gets the proposal}, the POT distance is small and correlates with the high value of AP. In contrast, when we replace the original query with an out-of-distribution one, the POT distance burgeons significantly, causing the AP to decrease from 90.28\% to 30.00\%. Therefore, we conclude that our READ framework is capable of intelligently adjusting the information flowing through the READ layers in order to produce the final output consistent with the video-language input and downstream tasks.

%% file: files/05_conclusion.tex
\section{Conclusion}
We propose a novel READ-PVLA framework for parameter-efficient transfer learning to video-language modeling tasks. Our READ-PVLA utilizes recurrent computation component to enable temporal modeling capability and partial video-language alignment objective to preserve critical information for bottleneck adaptation modules. Experiments demonstrate that READ-PVLA consistently outperforms both the full fine-tuning and competitive strategies, whilst bringing the benefit of parameter-efficiency (at most 1.20\% trainable parameters). Our method is also applicable to diverse pre-trained models, which has the potential to employ more powerful video-language models in the future.

%% file: files/06_ack.tex
\section*{Acknowledgements}
This research/project is supported by the National Research Foundation, Singapore under its AI Singapore Programme (AISG Award No: AISG3-PhD-2023-08-051T). Thong Nguyen is supported by a Google Ph.D. Fellowship in Natural Language Processing.